\documentclass[10pt,twocolumn,letterpaper]{article}

\usepackage{cvpr}              % To produce the CAMERA-READY version
\usepackage[accsupp]{axessibility} % Improves PDF readability for those with visual impairments.
\usepackage[dvipsnames]{xcolor}

\usepackage{graphicx}
\usepackage{amsmath}
\usepackage{amssymb}
\usepackage{booktabs}

\usepackage{bm}
\usepackage{multirow}
\usepackage{tabularx}
\usepackage{tabulary}
\usepackage{colortbl}
\usepackage{subcaption}

\usepackage[pagebackref, breaklinks=true, colorlinks, citecolor=citecolor, linkcolor=linkcolor, bookmarks=false]{hyperref}
\definecolor{citecolor}{HTML}{0071BC}
\definecolor{linkcolor}{HTML}{ED1C24}
\definecolor{namecolor}{HTML}{000000}
\definecolor{myblue}{HTML}{365694}

\newcommand{\RNum}[1]{\uppercase\expandafter{\romannumeral #1\relax}}

\newcommand{\highname}[1]{\textcolor{namecolor}{#1}}
\def\ours{\highname{GenTron}}

\def\paperID{862} % *** Enter the Paper ID here
\def\confName{CVPR}
\def\confYear{2024}

\title{
GenTron: Diffusion Transformers for Image and Video Generation
}

\author{
Shoufa Chen$^{1,2}$\textsuperscript{*} \quad Mengmeng Xu$^2$\textsuperscript{*} \quad Jiawei Ren$^2$ \quad Yuren Cong$^2$ \quad Sen He$^2$ \quad \\ Yanping Xie$^2$ \quad Animesh Sinha$^2$ \quad Ping Luo$^1$ \quad Tao Xiang$^2$ \quad Juan-Manuel Perez-Rua$^2$ \\
$^1$The University of Hong Kong \quad $^2$Meta
\vspace{-0.4cm}
}

\begin{document}
% \maketitle  % `maketitle` in teaser.tex

\twocolumn[{
\renewcommand\twocolumn[1][]{#1}
\maketitle
\begin{center}
    \centering
    \vspace*{-.15cm}
    \captionsetup{type=figure}\includegraphics[width=0.94\textwidth]{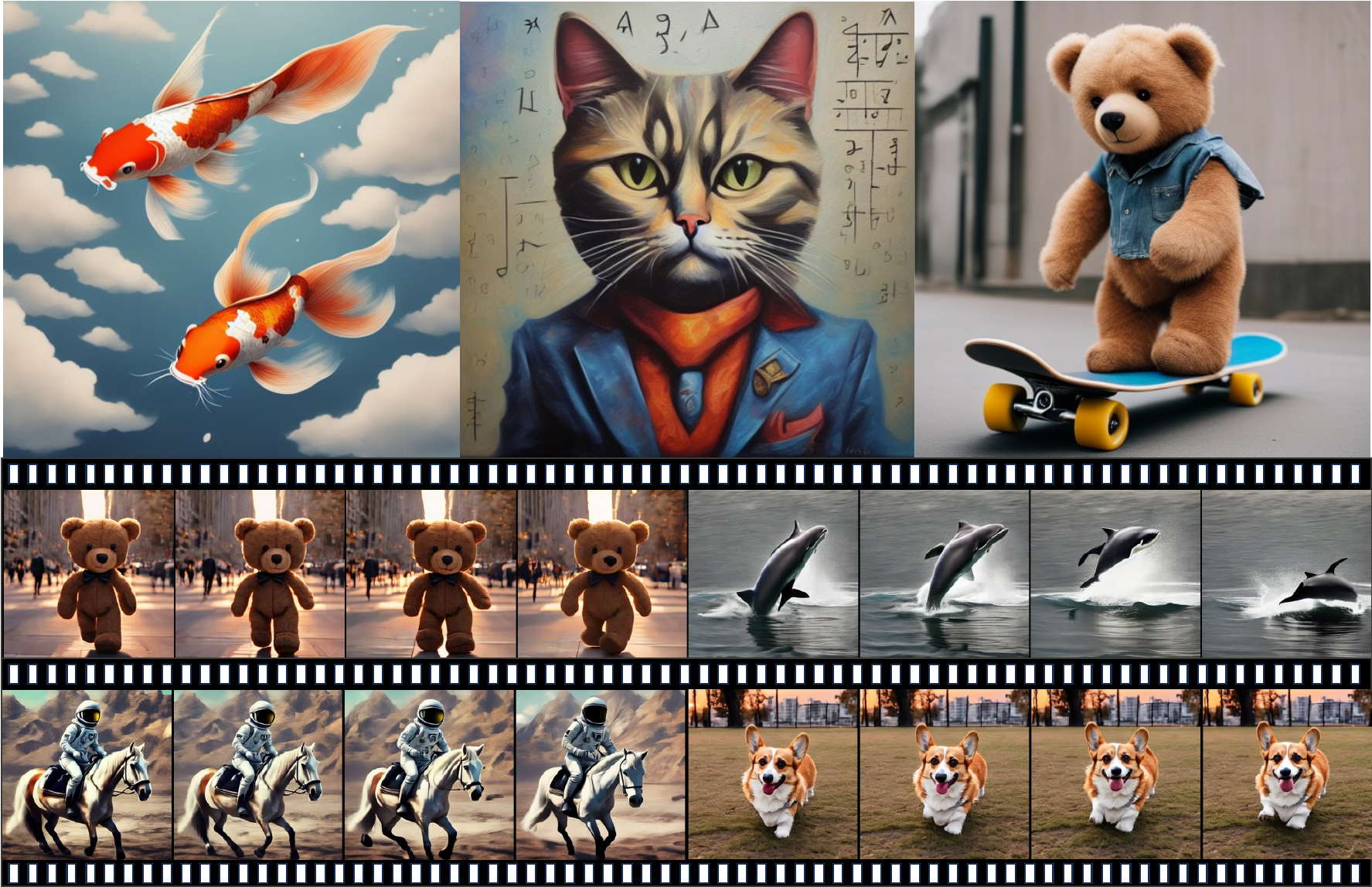}
    % \vspace*{.01cm}
    \captionof{figure}{\textbf{\ours{}: Transformer based diffusion model for high-quality text-to-image/video generation.}
    }
\label{fig:teaser}
\end{center}
}]

%%%%%%%%% ABSTRACT %%%%%%%%%

\begin{abstract}
In this study, we explore Transformer-based diffusion models for image and video generation.
Despite the dominance of Transformer architectures in various fields due to their flexibility and scalability, the visual generative domain primarily utilizes CNN-based U-Net architectures, particularly in diffusion-based models. 
We introduce \textbf{GenTron}, a family of \textbf{Gen}erative models employing \textbf{Tr}ansformer-based diffusi\textbf{on}, to address this gap.
Our initial step was to adapt Diffusion Transformers (DiTs) from class to text conditioning, a process involving thorough empirical exploration of the conditioning mechanism.
We then scale GenTron from approximately 900M to over 3B parameters, observing improvements in visual quality.
Furthermore, we extend GenTron to text-to-video generation, incorporating novel motion-free guidance to enhance video quality.
{\let\thefootnote\relax\footnote{{\textsuperscript{1}\url{https://www.shoufachen.com/gentron_website/}}}}
In human evaluations against SDXL, GenTron achieves a 51.1\% win rate in visual quality~(with a 19.8\% draw rate), and a 42.3\% win rate in text alignment~(with a 42.9\% draw rate). GenTron notably performs well in T2I-CompBench, highlighting its compositional generation ability. We hope GenTron could provide meaningful insights and serve as a valuable reference for future research. Please refer to the website\textsuperscript{1} and the arXiv version for the most up-to-date results: \url{https://arxiv.org/abs/2312.04557}.

\end{abstract}
%%%%%%%%% BODY TEXT
{\let\thefootnote\relax\footnote{{\textsuperscript{*}Equal contribution.}}}

\section{Introduction}

Diffusion models have recently shown remarkable progress in content creation, impacting areas such as image generation~\cite{ho2020denoising, saharia2022photorealistic, rombach2022high}, video generation~\cite{ho2022imagen, blattmann2023align, singer2022make, videoworldsimulators2024}, and editing~\cite{kawar2023imagic, cong2024flatten}. Among these notable developments, the CNN-based U-Net architecture has emerged as the predominant backbone design, a choice that stands in contrast to the prevailing trend in natural language processing~\cite{bert, brown2020language, touvron2023llama} and computer visual perception~\cite{dosovitskiy2021an, liu2021swin, Zhai_2022_CVPR} domains, where attention-based transformer architectures~\cite{vaswani2017attention} have empowered a renaissance and become increasingly dominant. To provide a comprehensive understanding of Transformers in diffusion generation and to bridge the gap in architectural choices between visual generation and the other two domains --- visual perception and NLP --- a thorough investigation of visual generation using Transformers is of substantial scientific value.

We focus on diffusion models with Transformers in this work. Specifically, our starting point is the foundational work known as DiT~\cite{peebles2022scalable}, which introduced a \emph{class}-conditioned latent diffusion model that employs a Transformer to replace the traditionally used U-Net architecture. We overcome the limitation of the original DiT model, which is constrained to handling only a restricted number~(\eg, 1000) of predefined classes, by utilizing language embeddings derived from open-world, free-form \emph{text} captions instead of predefined one-hot class embeddings. 
Along the way, we comprehensively investigate conditioning strategies, including (1) conditioning architectures: adaptive layer norm~(adaLN)~\cite{perez2018film} \vs cross-attention~\cite{vaswani2017attention}; and (2) text encoding methods: a generic large language model~\cite{chung2022scaling} \vs the language tower of multimodal models~\cite{radford2021learning}, or the combination of both of them. We additionally carry out comparative experiments and offer detailed empirical analyses to evaluate the effectiveness of these conditioning strategies.

Next, we explore the scaling-up properties of \ours{}. The Transformer architectures have been demonstrated to possess significant scalability in both visual perception~\cite{riquelme2021scaling, Zhai_2022_CVPR, chen2023pali, dehghani2023scaling} and language~\cite{radford2018improving, radford2019language, brown2020language, bert, touvron2023llama} tasks. For example, the largest dense language model has 540B parameters~\cite{chowdhery2022palm}, and the largest vision model has 22B~\cite{dehghani2023scaling} parameters. In contrast, the largest diffusion transformer, DiT-XL~\cite{peebles2022scalable}, only has about 675M parameters, trailed far behind both the Transformers utilized in other domains~(\eg, NLP) and recent diffusion arts with convolutional U-Net architectures~\cite{podell2023sdxl, dai2023emu}. To compensate for this considerable lagging, we scale up \ours{} in two dimensions, the number of transformer blocks and hidden dimension size, following the scaling strategy in~\cite{Zhai_2022_CVPR}. As a result, our largest model, \emph{\ours{}-G/2}, has more than 3B parameters and achieves significant visual quality improvement compared with the smaller one. 

Furthermore, we have advanced \ours{} from a T2I to a T2V model by inserting a temporal self-attention layer into each transformer block, making the first attempt to use transformers as the exclusive building block for video diffusion models. We also discuss existing challenges in video generation and introduce our solution, the \textbf{motion-free guidance~(MFG)}. Specifically, This approach involves intermittently \emph{disabling} motion modeling during training by setting the temporal self-attention mask to an identity matrix. Besides, MFG seamlessly integrates with the joint image-video strategy~\cite{dandi2020jointly, ho2022video, wang2023lavie, chen2023videocrafter1}, where images are used as training samples whenever motion is deactivated. Our experiments indicate that this approach clearly improves the visual quality of generated videos.

In human evaluations, GenTron outperforms SDXL, achieving a 51.1\% win rate in visual quality (with a 19.8\% draw rate), and a 42.3\% win rate in text alignment (with a 42.9\% draw rate). Furthermore, when compared to previous studies, particularly as benchmarked against T2I-CompBench~\cite{huang2023ticompbench}--a comprehensive framework for evaluating open-world compositional T2I generation--GenTron demonstrates superior performance across various criteria. These include attribute binding, object relationships, and handling of complex compositions.
 
Our \textbf{contributions} are summarized as follows:
(1) We have conducted a thorough and systematic investigation of transformer-based T2I generation with diffusion models. This study encompasses various conditioning choices and aspects of model scaling.
(2) In a pioneering effort, we explore a purely transformer-based diffusion model for T2V generation. We introduce \emph{motion-free guidance}, an innovative technique that efficiently fine-tunes T2I generation models for producing high-quality videos. (3) Experimental results indicate a clear preference for \ours{} over SDXL in human evaluations. Furthermore, \ours{} demonstrates superior performance compared to existing methods in the T2I-CompBench evaluations.

\section{Related Work}

\paragraph{Diffusion models for T2I and T2V generation.}
Diffusion models~\cite{ho2020denoising, nichol2021glide} are a type of generative model that creates data samples from random noise.
Later, latent diffusion models~\cite{rombach2022high, peebles2022scalable, podell2023sdxl} are proposed for efficient T2I generation. These designs usually have \emph{1)} a pre-trained Variational Autoencoder~\cite{kingma2013auto} that maps images to a compact latent space, \emph{2)} a conditioner modeled by cross-attention~\cite{rombach2022high, ho2022imagen} to process text as conditions with a strength control~\cite{ho2021classifierfree}, and \emph{3)} a backbone network, U-Net~\cite{ronneberger2015u} in particular, to process image features. 
The success of diffusion on T2I generation tasks underscores the promising potential for text-to-video~(T2V) generation~\cite{singer2022make, luo2023videofusion, khachatryan2023text2video, qi2023fatezero}. VDM~\cite{ho2022video} and Imagen Video~\cite{ho2022imagen} extend the image diffusion architecture on the temporal dimension with promising initial results.
To avoid excessive computing demands, video latent diffusion models~\cite{blattmann2023align, he2022latent, zhou2022magicvideo, wang2023modelscope, chen2023videocrafter1, lu2023vdt} implement the video diffusion process in a low-dimensional latent space.

\paragraph{Transformer-based Diffusion.}
Recently, Transformer-based Diffusion models have attracted increasing research interest. Among these, U-ViT~\cite{bao2022all} treats all inputs as tokens by integrating transformer blocks with a U-net architecture. In contrast, DiT~\cite{peebles2022scalable} employs a simpler, non-hierarchical transformer structure. MDT~\cite{gao2023masked} and MaskDiT~\cite{zheng2023fast} enhance DiT's training efficiency by incorporating the mask strategy~\cite{he2022masked}. Dolfin~\cite{wang2023dolfin} is a transformer-based model for layout generation. Concurrently to this work, PixArt-$\alpha$~\cite{chen2023pixart} demonstrates promising outcomes in Transformer-based T2I diffusion. It's trained using a three-stage decomposition process with high-quality data. Our work diverges from PixArt-$\alpha$ in key aspects. Firstly, while PixArt-$\alpha$ emphasizes training efficiency, our focus is on the design choice of conditioning strategy and scalability in T2I Transformer diffusion models. Secondly, we extend our exploration beyond image generation to video diffusion. We propose an innovative approach in video domain, which is not covered by PixArt-$\alpha$. 

\section{Method}
We first introduce the preliminaries in \Cref{sec:method-perliminary}, and then present the details of \ours{} for text-to-image generation in \Cref{sec:gentron-t2i}, which includes text encoder models, embedding integration methods, and scaling up strategy of \ours{}. Lastly, in \Cref{sec:gentron-t2v}, we extend GenTron's application to video generation, building on top of the T2I foundations laid in previous sections.

\subsection{Preliminaries}\label{sec:method-perliminary}

\paragraph{Diffusion models.}

Diffusion models~\citep{ho2020denoising} have emerged as a family of generative models that generate data by performing a series of transformations on random noise. 
They are characterized by a forward and a backward process.
Given an instance from the data distribution $x_0 \sim p(x_0)$, random Gaussian noise is iteratively added to the instance in the forward noising process to create a Markov Chain of random latent variable $x_1, x_2, ..., x_T$ following:
\begin{align}
    q(x_t|x_{t-1}) = \mathcal{N}(x_t;\sqrt{1-\beta_i}x_{t-1}, \beta_t\mathbf{I}),
\end{align} 
where $\beta_1, ... \beta_T$ are hyperparameters corresponding to the noise schedule. After a large enough number of diffusion steps, $x_T$ can be viewed as a standard Gaussian noise. 
A denoising network $\epsilon_\theta$ is further trained to learn the backward process, \ie, how to remove the noise from a noisy input~\citep{ho2020denoising}. 
For inference, an instance can be sampled starting from a random Gaussian noise $x_T\sim\mathcal{N}(0;\mathbf{I})$ and denoised step-by-step following the Markov Chain, \ie, by sequentially sampling $x_{t-1}$ to $x_0$ with  $p_\theta(x_{t-1}|x_t)$:
\begin{align}
\label{eq:diff_sample}
 x_{t-1} = \frac{1}{\sqrt{\alpha_t}} \Bigl(x_t - \frac{1-\alpha_t}{\sqrt{1-\bar{\alpha}_t}}\epsilon_\theta(x_t, t)\Bigl) + \sigma_t\mathbf{z},
\end{align}
where $\bar{\alpha}_t=\prod_{s=1}^t\alpha_s$, $\alpha_t=1-\beta_t$ and $\sigma_t$ is the noise scale.
In practical application, the diffusion sampling process can be further accelerated using different sampling techniques~\citep{song2020denoising, lu2022dpm}. 

\begin{figure}[t]
    \centering
    \begin{subfigure}[b]{0.4\linewidth}
    \includegraphics[width=\textwidth]{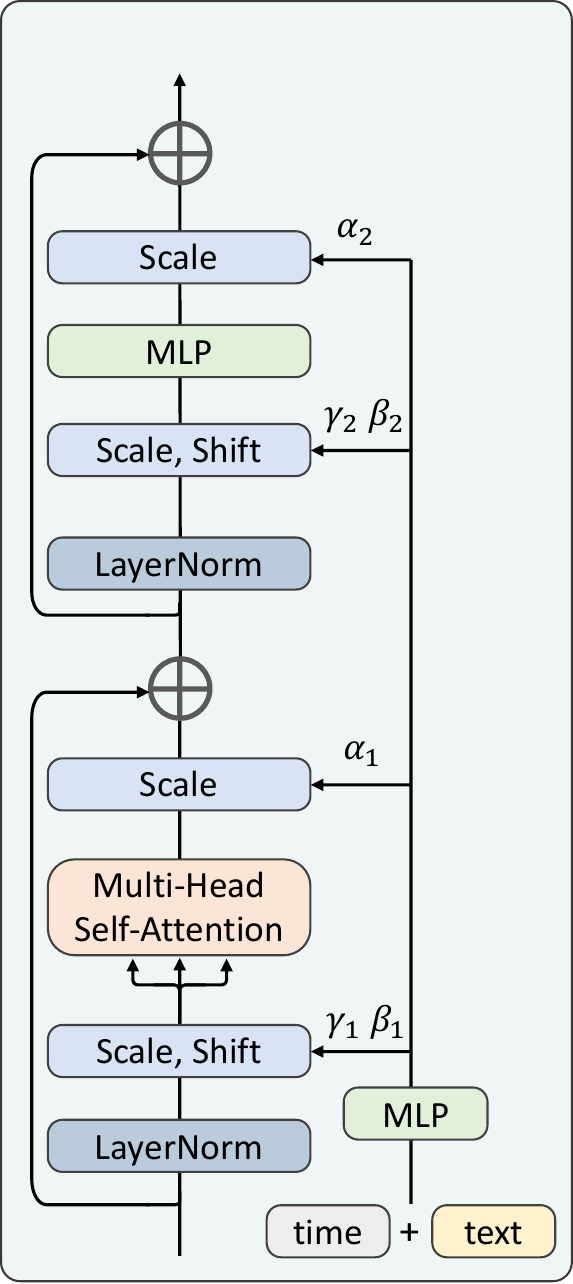}
    \caption{adaLN-Zero}
    \label{fig:arch_adaln}
    \end{subfigure}
    \hspace{0.06\linewidth}
    \begin{subfigure}[b]{0.4\linewidth}
    \includegraphics[width=\textwidth]{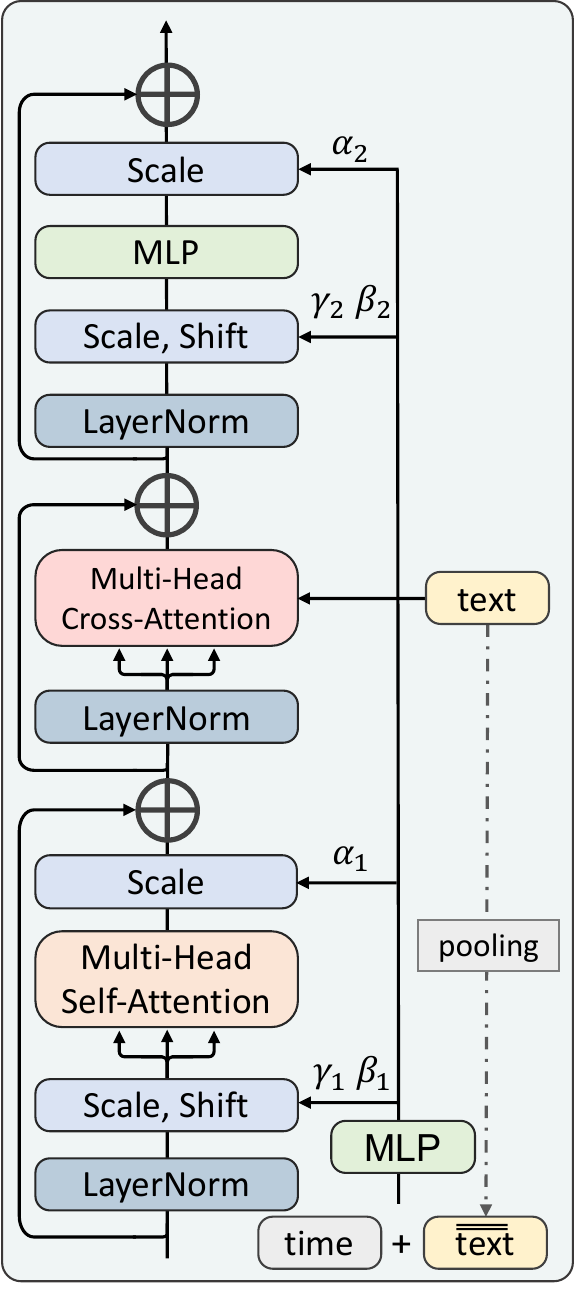}
    \caption{Cross attention}\label{fig:arch_crossattn}
    \end{subfigure}
    \vspace{-4pt}
    \caption{\textbf{Text embedding integration architecture.} We directly adapt adaLN from DiT~\cite{peebles2022scalable}, substituting the one-hot \emph{class} embedding with \emph{text} embedding. For cross attention, different from the approach in~\cite{peebles2022scalable}, we maintain the use of adaLN to model the combination of \emph{time} embedding and the aggregated \emph{text} embedding.}
    \label{fig:adaLN-vs-crossattn-arch}
    \vspace{-6pt}
\end{figure}

\paragraph{Latent diffusion model architectures.}
Latent diffusion models~(LDMs)~\cite{rombach2022high} reduce the high computational cost by conducting the diffusion process in the latent space. First, a pre-trained autoencoder~\cite{kingma2013auto, esser2021taming} is utilized to compress the raw image from pixel to latent space, then the diffusion models, which are commonly implemented with a U-Net~\cite{ronneberger2015u} backbone, work on the latent space. Peebles~\etal proposed DiT~\cite{peebles2022scalable} to leverage the transformer architecture as an alternative to the traditional U-Net backbone for \emph{class}-conditioned image generation, adopting the adaptive layernorm~(adaLN~\cite{perez2018film}) for class conditioning mechanism, as shown in~\Cref{fig:arch_adaln}.

\subsection{Text-to-Image \ours{}}\label{sec:gentron-t2i}

Our \ours{} is built upon the DiT-XL/2~\cite{peebles2022scalable}, which converts the latent of shape 32$\times$32$\times$4 to a sequence of non-overlapping tokens with a 2$\times$2 patchify layer~\cite{dosovitskiy2021an}. Then, these tokens are sent into a series of transformer blocks. Finally, a standard linear decoder is applied to convert these image tokens into latent space.

While DiT has shown that transformer-based models yield promising results in \emph{class}-conditioned scenarios, it did not explore the realm of T2I generation. This field poses a considerable challenge, given its less constrained conditioning format. Moreover, even the largest DiT model, DiT-XL/2, with its 675 million parameters, is significantly overshadowed by current U-Nets~\cite{podell2023sdxl, dai2023emu}, which boast over 3 billion parameters. To address these limitations, our research conducts a thorough investigation of transformer-based T2I diffusion models, focusing specifically on text conditioning approaches and assessing the scalability of the transformer architecture by expanding \ours{} to more than 3 billion parameters.

\subsubsection{From \emph{Class} to \emph{Text} Condition}
T2I diffusion models rely on textual inputs to steer the process of image generation. The mechanism of text conditioning involves two critical components: firstly, the selection of a text encoder, which is responsible for converting raw text into text embeddings, and secondly, the method of integrating these embeddings into the diffusion process.  For a complete understanding, we have included in the appendix a detailed presentation of the decisions made in existing works concerning these two components.

\paragraph{Text encoder model.} 
Current advancements in T2I diffusion techniques employ a variety of language models, each with its unique strengths and limitations. To thoroughly assess which model best complements transformer-based diffusion methods, we have integrated several models into GenTron. This includes the text towers from multimodal models, CLIP~\cite{radford2021learning}, as well as a pure large language model, Flan-T5~\cite{chung2022scaling}. Our approach explores the effectiveness of these language models by integrating each model independently with GenTron to evaluate their individual performance and combinations of them to assess the potential properties they may offer when used together.

\paragraph{Embedding integration.}
In our study, we focused on two methods of embedding integration: adaptive layernorm and cross-attention.
(1) \textbf{Adaptive layernorm~(adaLN).} As shown in \Cref{fig:arch_adaln}, this method integrates conditioning embeddings as normalization parameters on the feature channel. Widely used in conditional generative modeling, such as in StyleGAN~\cite{karras2019style}, adaLN serves as the standard approach in DiT~\cite{peebles2022scalable} for managing class conditions.
(2) \textbf{Cross-attention.} As illustrated in \Cref{fig:arch_crossattn}, the image feature acts as the \texttt{query}, with textual embedding serving as \texttt{key} and \texttt{value}. This setup allows for direct interaction between the image feature and textual embedding through an attention mechanism~\cite{vaswani2017attention}. Besides, different from the cross-attention discussed in~\cite{peebles2022scalable}, which processes the class embedding and time embedding together by firstly concatenating them, we maintain the use of adaLN in conjunction with the cross-attention to separately model the \emph{time} embedding. The underlying rationale for this design is our belief that the time embedding, which is consistent across all spatial positions, benefits from the global modulation capabilities of adaLN. Moreover, we also add the pooled text embeddings to the time embedding followings~\cite{podell2023sdxl, balaji2022ediffi, ho2022cascaded, nichol22aglide}.

\subsubsection{Scaling Up \ours{}}

\begin{table}[t]
    \small
    \centering
    \begin{tabular}{l c c c c}
    Model & Depth & Width & MLP Width & \#Param.  \\
    \hline 
    \ours{}-XL/2  & 28 & 1152 & 4608 & 930.0M\\
    \ours{}-G/2   & 48 & 1664 & 6656 & 3083.8M\\
    \end{tabular}
    \vspace{-2pt}
    \caption{\textbf{Configuration details of \ours{} models}.}
    \label{tab:arch_scaleup}
    \vspace{-6pt}
\end{table}

To explore the impact of substantially scaling up the model size, we have developed an advanced version of \ours{}, which we refer to as \ours{}-G/2. This model was constructed in accordance with the scaling principles outlined in~\cite{Zhai_2022_CVPR}. We focused on expanding three critical aspects: the number of transformer blocks~(\emph{depth}), the dimensionality of patch embeddings~(\emph{width}), and the hidden dimension of the MLP~(\emph{MLP-width}). The specifications and configurations of the GenTron models are detailed in \Cref{tab:arch_scaleup}. Significantly, the GenTron-G/2 model boasts over 3 billion parameters. To our knowledge, this represents the largest transformer-based diffusion architecture developed to date.

\subsection{Text-to-Video \ours{}}\label{sec:gentron-t2v}

In this subsection, we elaborate on the process of adapting \ours{} from a T2I framework to a T2V framework. \cref{sec:t2v-arch} will detail the modifications made to the model's architecture, enabling \ours{} to process video data. Furthermore, \cref{sec:t2v-mfg} will discuss the challenges encountered in the domain of video generation and the innovative solutions we have proposed to address them.

\subsubsection{\ours{}-T2V Architecture}\label{sec:t2v-arch}

\begin{figure}[t]
    \centering
    \includegraphics[width=0.7\linewidth]{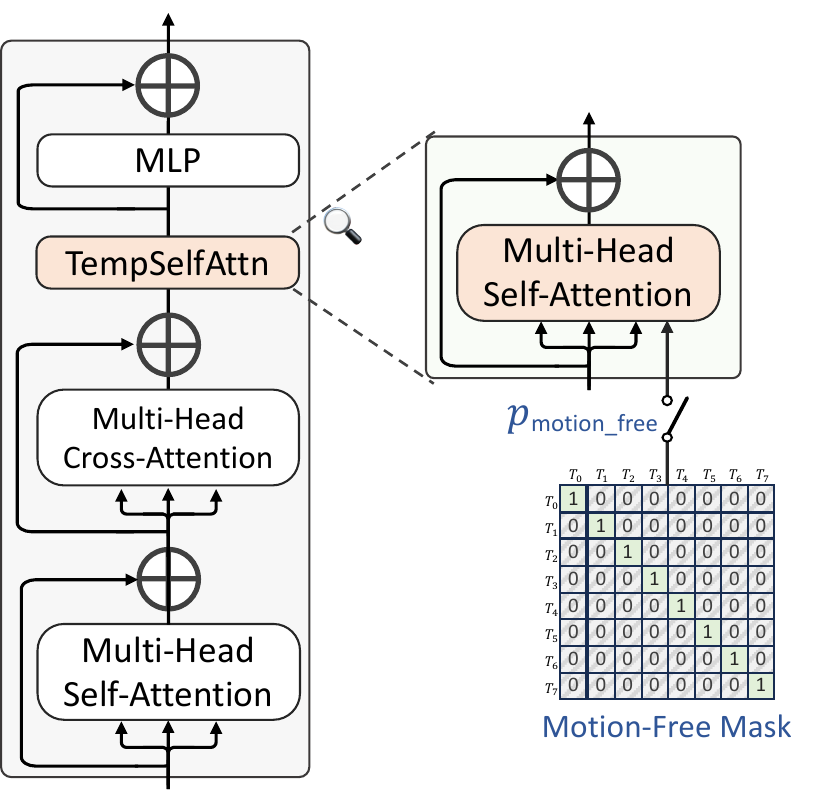}
    \caption{\textbf{\ours{} Transformer block with $\operatorname{TempSelfAttn}$ and Motion-Free Mask.} The temporal self-attention layer is inserted between the cross-attention and the MLPs. The motion-free mask, which is an identity matrix, will be utilized in the $\operatorname{TempSelfAttn}$ with a probability of \textcolor{myblue}{$p_{\text{motion\_free}}$}. We omit details like text conditioning, $\operatorname{LN}$ here for simplicity, which could be found in \Cref{fig:adaLN-vs-crossattn-arch}.}\label{fig:t2v-block-arch}
    \vspace{-8pt}
\end{figure}

\paragraph{Transformer block with temporal self-attention.}
It is typically a common practice to train video diffusion models from image diffusion models by adding new temporal modeling modules~\cite{ho2022video, blattmann2023align, zhou2022magicvideo, guo2023animatediff}. These usually consist of 3D convolutional layers and temporal transformer blocks that focus on calculating attention along the temporal dimension. 
In contrast to the traditional approach~\cite{blattmann2023align}, which involves adding both temporal convolution layers and temporal transformer blocks to the T2I U-Net, our method integrates only lightweight temporal self-attention~($\operatorname{TempSelfAttn}$) layers into each transformer block. As depicted in \Cref{fig:t2v-block-arch}, the $\operatorname{TempSelfAttn}$ layer is placed right after the cross-attention layer and before the MLP layer. Additionally, we modify the output of the cross-attention layer by reshaping it before it enters the $\operatorname{TempSelfAttn}$ layer and then reshape it back to its original format once it has passed through. This process can be formally represented as:
\begin{align}
    \mathbf{x} &= \texttt{rearrange}(\mathbf{x}, \texttt{(b t) n d} \xrightarrow{} \texttt{(b n) t d} ) \\
    \mathbf{x} &= \mathbf{x} + \operatorname{TempSelfAttn}(\operatorname{LN}(\mathbf{x})) \label{eq:temp-self-attn} \\
    \mathbf{x} &= \texttt{rearrange}(\mathbf{x}, \texttt{(b n) t d} \xrightarrow{} \texttt{(b t) n d})
\end{align}
\noindent where \texttt{b}, \texttt{t}, \texttt{n}, \texttt{d} represent the batch size, number of frames, number of patches per frame, and channel dimension, respectively. \texttt{rearrage} is a notation from~\cite{rogozhnikov2021einops}. We discovered that a simple $\operatorname{TempSelfAttn}$ layer suffices to capture motion, a finding that aligns with observations in a recent study~\cite{wang2023lavie}. In addition, only using $\operatorname{TempSelfAttn}$ makes it convenient to \emph{turn on} and \emph{turn off} the temporal modeling, which would be discussed in \cref{sec:t2v-mfg}. 

\paragraph{Initialization.}
We use the pre-trained T2I model as a basis for initializing the shared layers between T2I and T2V models. In addition, for the newly added $\operatorname{TempSelfAttn}$ layers, we initialize the weights and biases of the output project layers to zero. This ensures that at the beginning of the T2V fine-tuning stage, these layers produce a zero output, effectively functioning as an identity mapping in conjunction with the shortcut connection.

\begin{table*}[t]
\centering
\setlength{\tabcolsep}{4pt}
\renewcommand{\arraystretch}{1}
\small
\begin{tabular}{lccccccccccc}
\toprule
\multicolumn{3}{c}{Conditioning} &  \multirow{2}{*}{Scale} & \multicolumn{3}{c}{Attribute Binding } & \multicolumn{2}{c}{ Object Relationship} & \multirow{2}{*}{Complex} & \multirow{2}{*}{\textbf{Mean}}
\\
\cmidrule(lr){1-3}\cmidrule(lr){5-7}\cmidrule(lr){8-9}
Text Encoder & Type &  Integration & & 
Color &
Shape &
Texture &
Spatial &
Non-spatial &
\\
\midrule
CLIP-L~\cite{radford2021learning} & MM & adaLN-zero & XL/2 & 36.94 & 42.06 & 50.73 & 9.41 & 30.38 & 36.41 & 34.32 \\
CLIP-L~\cite{radford2021learning} & MM & cross-attn & XL/2  & 73.91 & 51.81 & 68.76 & 19.26 & 31.80 & 41.52 & 47.84 \\
T5-XXL~\cite{chung2022scaling} & LLM & cross-attn & XL/2 & 74.90 & 55.40 & 70.05 & 20.52 & 31.68 & 41.01 & 48.93 \\
CLIP-T5XXL & MM + LLM & cross-attn & XL/2 & 75.65 & 55.74 & 69.48 & 20.67 & 31.79 & 41.44 & 49.13 \\
CLIP-T5XXL & MM + LLM & cross-attn & G/2 & \bf{76.74} & \bf{57.00} & \bf{71.50}  & \bf{20.98} & \bf{32.02} & \bf{41.67} &  \bf{49.99} \\ 
\bottomrule
\end{tabular}
\vspace{-4pt}
\caption{\textbf{Conditioning and model scale in \ours{}.} We compare \ours{} model variants with different design choices on T2I-CompBench~\cite{huang2023ticompbench}. The text encoders are from the language tower of the multi-modal~(MM) model, the large language model~(LLM), or a combination of them. The \ours{}-G/2 with CLIP-T5XXL performs best. Detailed discussions can be found in \cref{sec:result-t2i}.}\label{tab:ablation-condition}
\vspace{-6pt}
\end{table*}

\subsubsection{Motion-Free Guidance}\label{sec:t2v-mfg}

\paragraph{Challenges encountered.}
We observed a notable phenomenon in the current T2V diffusion models~\cite{ho2022video, blattmann2023align} where the per-frame visual quality significantly lags behind that of T2I models~\cite{rombach2022high, xue2023raphael, podell2023sdxl, dai2023emu}. Furthermore, our analysis revealed a remarkable degradation in visual quality in the T2V models post-fine-tuning, especially when compared to their original T2I counterparts. 
We note that these problems generally exist in current T2V diffusion models, not limited to our transformer-based T2V.

\paragraph{Problem analysis and insights.}
We presume that the observed lag in the visual quality of T2V primarily stems from two factors: the nature of video data and the fine-tuning approach. Firstly, publicly available video datasets often fall short in both quality and quantity compared to image datasets. For instance, \cite{schuhmann2022laion} has more than 2B English image-text pairs, whereas the current widely used video dataset, WebVid-10M~\cite{bain2021frozen} contains only 10.7M video-text pairs. Additionally, many video frames are compromised by motion blur and watermarks, further reducing their visual quality. This limited availability hampers the development of robust and versatile video diffusion models.
Secondly, the focus on optimizing temporal aspects during video fine-tuning can inadvertently compromise the spatial visual quality, resulting in a decline in the overall quality of the generated videos.

\paragraph{Solution \RNum{1}: joint image-video training.}
From the data aspect, we adopt the joint image-video training strategy~\cite{dandi2020jointly, ho2022video, wang2023lavie, chen2023videocrafter1} to mitigate the video data shortages.
Furthermore, joint training helps to alleviate the problem of domain discrepancy between video and image datasets by integrating both data types for training.

\paragraph{Solution \RNum{2}: motion-free guidance.} We treat the temporal motion within a video clip as a special \emph{conditioning} signal, which can be analogized to the textual conditioning in T2I/T2V diffusion models. Based on this analogy, we propose a novel approach, \emph{motion-free guidance}~(MFG), inspired by classifier-free guidance~\cite{ho2021classifierfree, brooks2023instructpix2pix}, to modulate the weight of motion information in the generated video.

In a particular training iteration, our approach mirrors the concept used in classifier-free guidance, where conditioned text is replaced with an empty string. The difference is that we employ an identity matrix to \emph{nullify} the temporal attention with a probability of \textcolor{myblue}{$p_{\text{motion\_free}}$}. This identity matrix, which is depicted in \Cref{fig:t2v-block-arch}~(\textcolor{myblue}{Motion-Free Mask}), is structured such that its diagonal is populated with ones, while all other positions are zeroes. This configuration confines the temporal self-attention to work within a single frame. Furthermore, as introduced in \cref{sec:t2v-arch}, temporal self-attention is the sole operator for temporal modeling. Thus, using a motion-free attention mask suffices to disable temporal modeling in the video diffusion process.

During inference, we have text and motion conditionings. Inspired by~\cite{brooks2023instructpix2pix}, we can modify the score estimate as:
\begin{align}
    \tilde{\epsilon}_\theta &= \epsilon_\theta(x_t, \varnothing, \varnothing) \nonumber \\
    &+ \lambda_T \cdot (\epsilon_\theta(x_t, c_T, c_M) - \epsilon_\theta(x_t, \varnothing, c_M)) \nonumber \\
    &+ \lambda_M \cdot (\epsilon_\theta(x_t, \varnothing, c_M) - \epsilon_\theta(x_t, \varnothing, \varnothing))
\end{align}
where $c_T$ and $c_M$ represent the text conditioning and motion conditioning. $\lambda_T$ and $\lambda_M$ are the guidance scale of standard text and that of motion, controlling how strongly the generated samples correspond with the text condition and the motion strength, respectively. 
We empirically found that fixing $\lambda_T = 7.5$ and adjusting $\lambda_M \in [1.0, 1.3]$ for each example tend to achieve the best result. This finding is similar to~\cite{brooks2023instructpix2pix}, although our study utilizes a narrower range for $\lambda_M$.

\paragraph{Putting solutions together.} We can integrate solution \RNum{1} and \RNum{2} together in the following way: when the motion is omitted at a training step, we load an image-text pair and repeat the image $T - 1$ times to create a \emph{pseudo} video. Conversely, if motion is included, we instead load a video clip and extract it into $T$ frames.

\begin{table}[t]
\centering
\setlength{\tabcolsep}{2 pt}
\renewcommand{\arraystretch}{0.9}
\footnotesize
\begin{tabular}{lccccccc}
\toprule
\multicolumn{1}{c}
{\multirow{2}{*}{Model}} & \multicolumn{3}{c}{Attribute Binding } & \multicolumn{2}{c}{Obj. Relation} & \multirow{2}{*}{Comp.} & \multirow{2}{*}{\textbf{Mean}}
\\
\cmidrule(lr){2-4}\cmidrule(lr){5-6}
&
Color &
Shape &
Texture &
Spat. &
Non-spat. &
\\
\midrule
% convert 0.xxx to percent format
% fn = lambda x: " & ".join(["{:.2f}".format(100 * float(xx)) for xx in x.split("&") ])
LDM v1.4 & 37.65 & 35.76 & 41.56 & 12.46 & 30.79 & 30.80 & 31.50 \\
LDM v2   & 50.65 & 42.21 & 49.22 & 13.42 & 30.96 & 33.86 & 36.72 \\
Composable v2 & 40.63 & 32.99 & 36.45 & 8.00 & 29.80 & 28.98 & 29.47 \\
Structured v2 & 49.90 & 42.18 & 49.00 & 13.86 & 31.11 & 33.55 & 36.60 \\
Attn-Exct v2 & 64.00 & 45.17 & 59.63 & 14.55 & 31.09 & 34.01 & 41.41 \\
GORS  & 66.03 & 47.85 & 62.87 & 18.15 & 31.93 & 33.28 & 43.35 \\
DALL·E 2 & 57.50 & 54.64 & 63.74 & 12.83 & 30.43 & 36.96 & 42.68 \\
LDM XL & 63.69 & 54.08 & 56.37 & 20.32 & 31.10 & 40.91 & 44.41 \\
PixArt-$\alpha$ & 68.86 & 55.82 & 70.44 & 20.82 & 31.79 & 41.17 & 48.15 \\
\midrule
\ours{} & \bf{76.74}	& \bf{57.00} & \bf{71.50}  & \bf{20.98} & \bf{32.02} & \bf{41.67}  & \bf{49.99} \\ 
\bottomrule
\end{tabular}
\vspace{-4pt}
\caption{\textbf{Comparison of alignment evaluation on T2I-CompBench~\cite{huang2023ticompbench}.} Results show our advanced model, \ours{}{\footnotesize-CLIPT5XXL-G/2}, achieves superior performance across multiple compositional metrics compared to previous methods.
}\label{benchmark:t2icompbench}
\vspace{-6pt}
\end{table}

\section{Experiments}

\subsection{Implementation Details}
\paragraph{Training scheme} 
For all \ours{} model variations, we employ the AdamW~\cite{loshchilov2017decoupled} optimizer, maintaining a constant learning rate of 1$\times$10$^{-4}$. 
We train our T2I \ours{} models in a multi-stage procedure~\cite{rombach2022high, podell2023sdxl} with an internal dataset, including a low-resolution~(256$\times$256) training with a batch size of 2048 and 500K optimization steps, as well as high-resolution~(512$\times$512) with a batch size of 784 and 300K steps. For the \ours{}-G/2 model, we further integrate Fully Sharded Data Parallel~(FSDP)~\cite{zhao2023pytorch} and activation checkpointing~(AC), strategies specifically adopted to optimize GPU memory usage. In our video experiments, we train videos on a video dataset that comprises approximately 34M videos. To optimize storage usage and enhance data loading efficiency, the videos are pre-processed to a resolution with a short side of 512 pixels and a frame rate of 24 FPS. We process batches of 128 video clips. Each clip comprises 8 frames, captured at a sampling rate of 4 FPS.

\paragraph{Evaluation metrics.} We mainly adopt the recent T2I-CompBench~\cite{huang2023ticompbench} to compare \ours{} model variants, following~\cite{chen2023pixart, dalle3}. Specifically, we compare the attribute binding aspects, which include \emph{color}, \emph{shape}, and \emph{texture}. We also compare the spatial and non-spatial object relationships. Moreover, user studies are conducted to compare visual quality and text alignment.

\subsection{Main Results of \ours{}-T2I}\label{sec:result-t2i}
In this subsection, we discuss our experimental results,
focusing on how various conditioning factors and model sizes impact \ours{}'s performance. \Cref{tab:ablation-condition} presents the quantitative findings of our study. Additionally, we offer a comparative visualization, illustrating the effects of each conditioning factor we explored. A comparison to prior art is also provided.

\begin{figure}[t]
    \centering
    \begin{subfigure}[b]{0.47\linewidth}
    \includegraphics[width=\textwidth]{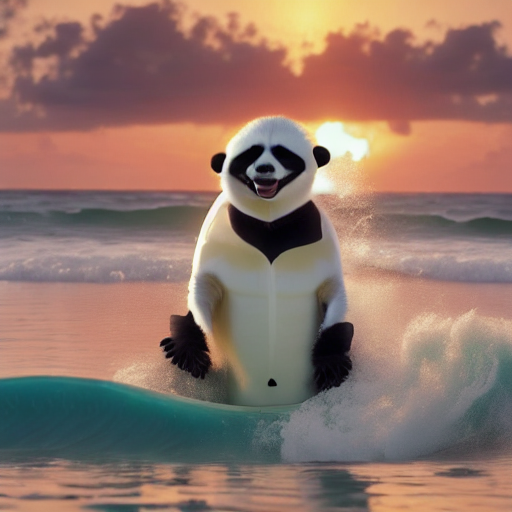}
    \caption{adaLN-Zero}
    \end{subfigure}
    \begin{subfigure}[b]{0.47\linewidth}
    \includegraphics[width=\textwidth]{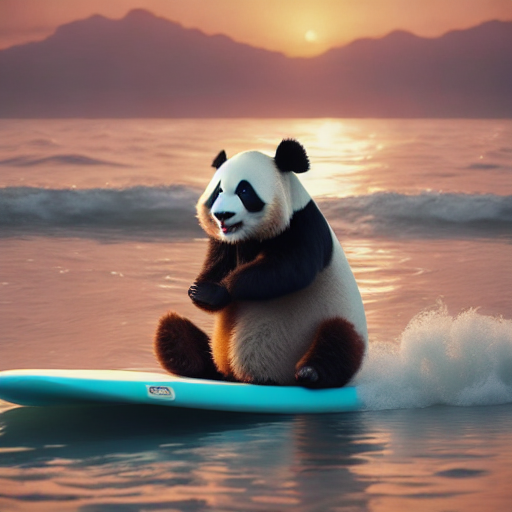}
    \caption{Cross attention}
    \end{subfigure}
    \vspace{-6pt}
    \caption{\textbf{adaLN-Zero \vs cross attention.} The prompt is \textit{\footnotesize ``A panda standing on a surfboard in the ocean in sunset.''} Cross attention exhibits a distinct advantage in the \emph{text}-conditioned scenario.}
    \label{fig:adaLN-vs-crossattn}
    \vspace{-8pt}
\end{figure}
\paragraph{Cross attention \vs adaLN-Zero.}
Recent findings~\cite{peebles2022scalable} conclude that the adaLN design yields superior results in terms of the FID, outperforming both cross-attention and in-context conditioning in efficiency for \emph{class}-based scenarios. However, our observations reveal a limitation of adaLN in handling free-form text conditioning. This shortcoming is evident in \Cref{fig:adaLN-vs-crossattn}, where adaLN's attempt to generate a panda image falls short, with cross-attention demonstrating a clear advantage. This is further verified quantitatively in the first two rows of \Cref{tab:ablation-condition}, where cross-attention uniformly excels over adaLN in all evaluated metrics. 

This outcome is reasonable considering the nature of \emph{class} conditioning, which typically involves a limited set of fixed signals~(\eg, the 1000 one-hot class embeddings for ImageNet~\cite{deng2009imagenet}). In such contexts, the adaLN approach, operating at a spatially global level, adjusts the image features uniformly across all positions through the normalization layer, making it adequate for the static signals. In contrast, cross-attention treats spatial positions with more granularity. It differentiates between various spatial locations by dynamically modulating image features based on the cross-attention map between the text embedding and the image features. This spatial-sensitive processing is essential for free-from \emph{text} conditioning, where the conditioning signals are infinitely diverse and demand detailed representation in line with the specific content of textual descriptions.

\paragraph{Comparative analysis of text encoders.} In \Cref{tab:ablation-condition}~(rows two to four), we conduct a quantitative evaluation of various text encoders on T2I-CompBench, ensuring a fair comparison by maintaining a consistent XL/2 size across models. Results reveal that \ours{}-T5XXL outperforms \ours{}-CLIP-L across all three attribute binding and spatial relationship metrics, while it demonstrates comparable performance in the remaining two metrics. This suggests that T5 embeddings are superior in terms of compositional ability. These observations are in line with~\cite{balaji2022ediffi}, which utilizes both CLIP and T5 embeddings for training but tests them individually or in combination during inference. Unlike eDiff, our approach maintains the same settings for both training and inference. Notably, \ours{} demonstrates enhanced performance when combining CLIP-L and T5XXL embeddings, indicating the model's ability to leverage the distinct advantages of each text embedding type.

\paragraph{Scaling \ours{} up.} 
\begin{figure}[t]
    \centering
    \begin{subfigure}[b]{0.47\linewidth}
    \includegraphics[width=\textwidth]{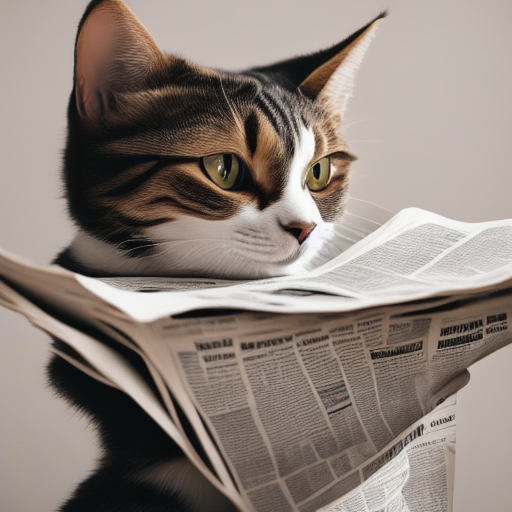}
    \caption{\ours{}-XL/2}
    \end{subfigure}
    \begin{subfigure}[b]{0.47\linewidth}
    \includegraphics[width=\textwidth]{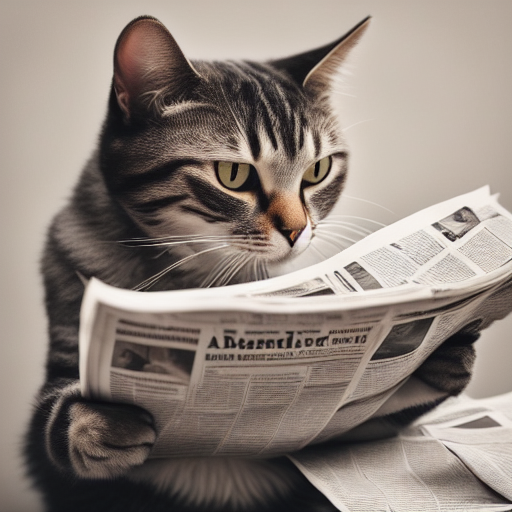}
    \caption{\ours{}-G/2}
    \end{subfigure} 
    \vspace{-6pt}
    \caption{\textbf{Effect of model scale.} The prompt is {\footnotesize``a cat reading a newspaper''}. The larger model \ours{}-G/2 excels in rendering finer details and rationalization in the layout of the cat and newspaper. 
    More comparisons can be found in the appendix.
    }
    \label{fig:scaleup}
    \vspace{-8pt}
\end{figure}

In \Cref{fig:scaleup}, we showcase examples from the PartiPrompts benchmark~\cite{yu2022scaling} to illustrate the qualitative enhancements achieved by scaling our model up from approximately 900 million to 3 billion parameters. Both models operate under the same CLIP-T5XXL condition. 
\begin{figure*}[t]
    \centering
    \includegraphics[width=0.96\linewidth]{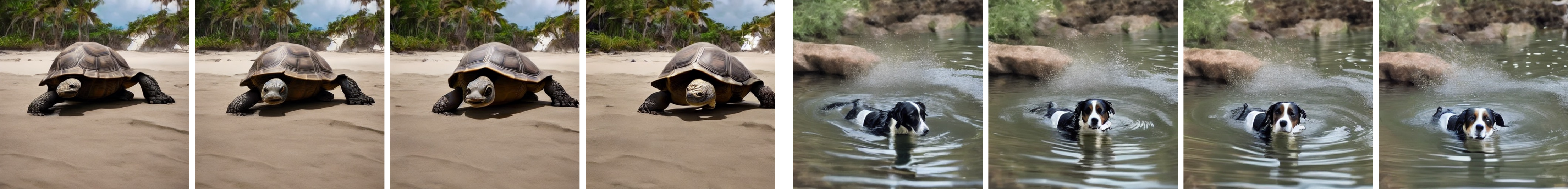}
    \vspace{-4pt}
    \caption{\textbf{\ours{}-T2V examples.} Prompts are ``A giant tortoise is making its way across the beach'' and ``A dog swimming''.}.\label{fig:vis_t2v}
    \vspace{-8pt}
\end{figure*}
The larger \ours{}-G/2 model excels in rendering finer details and more accurate representations, particularly in rationalizing the layout of objects like cats and newspapers. This results in image compositions that are both more coherent and more realistic. In comparison, the smaller \ours{}-XL/2 model, while producing recognizable images with similar color schemes, falls short in terms of precision and visual appeal.

Furthermore, the superiority of \ours{}-G/2 is quantitatively affirmed through its performance on T2I-CompBench, as detailed in \Cref{tab:ablation-condition}. The increase in model size correlates with significant improvements across all evaluative criteria for object composition, including attributes and relationships. Additional comparative examples are provided in the appendix for further illustration.

\begin{table}[t]
    \setlength{\tabcolsep}{1pt}
    \renewcommand{\arraystretch}{1.0}
    \small
    \begin{center}
        \begin{tabular}{lcccc}
    Method       & \#Param. & FID-30K\textcolor{cyan}{$\downarrow$}  & CLIP-Score\textcolor{cyan}{$\uparrow$}  & T2I-CompBen\textcolor{cyan}{$\uparrow$} \\
    \hline
    Imagen       & 3.0B     & \textbf{7.27}        & 0.27                 & - \\
    Parti-750M   & 0.8B     & 10.71                & -                    & -   \\
    Parti-3B     & 3.0B     & 8.10                 & -                    & -  \\
    GigaGAN      & 1.0B     & 9.09                 & 0.322                & -    \\
    MUSE-3B      & 3.0B     & 7.88                 & 0.320                & -      \\
    SD v1.4      & 0.9B     & 12.94                & 0.325                & 31.50    \\
    SDXL         & 2.6B     & 17.82	               & 0.329                & 44.41   \\
    \hline
    GenTron-XL/2 & 0.9B     & 14.21        & 0.326                & 49.13   \\
    GenTron-G/2  & 3.1B     & 14.53        & \textbf{0.335}       & \textbf{49.99}    \\
    \end{tabular}
    \caption{\textbf{Comparison to T2I models.}}\label{tab:t2ifid}
    \end{center}
    \vspace{-15pt}
\end{table}

\paragraph{Comparison to prior work.}
In \Cref{benchmark:t2icompbench}, we showcase the alignment evaluation results from T2I-CompBench.  Our method demonstrates outstanding performance in all areas, including attribute binding, object relationships, and complex compositions. This indicates a heightened proficiency in compositional generation, with a notable strength in color binding. In this aspect, our approach surpasses the previous state-of-the-art (SoTA) work~\cite{chen2023pixart} by over 7\%.

In \Cref{tab:t2ifid}, we further thoroughly compare the zero-shot FID-30K, CLIP-Score, and T2I-CompBench metrics with previous T2I models.
In contrast to U-Net based SDv1.4, GenTron uses about \emph{four times \textbf{less} data} (550M \vs 2B), while achieving better CLIP-scores and T2I-CompBench results. Although GenTron does not achieve a superior FID score compared to SDv1.4, it's important to note that recent studies, including Pick-a-Pic~\cite{kirstain2024pick} and SDXL~\cite{podell2023sdxl}, have highlighted that FID scores often misrepresent human preferences in generative models, sometimes even \textbf{negatively} correlating with visual aesthetics.

\begin{figure}[t]
    \centering
    \includegraphics[width=0.96\linewidth]{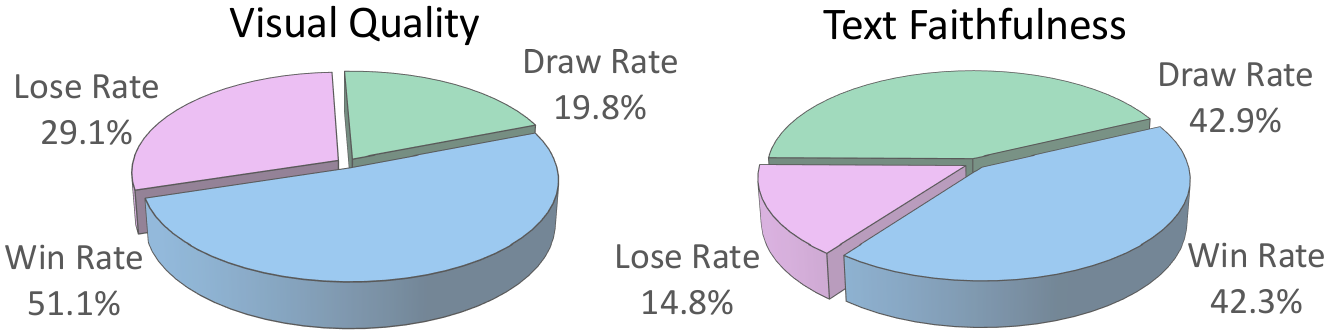}
    \vspace{-1pt}
    \caption{\textbf{Visualization of the human preference of our method \vs Latent Diffusion XL.} Our method received a significantly higher number of votes as the winner in comparisons of visual quality and text faithfulness, with a total of 3000 answers.}
    \label{fig:userstudy}
    \vspace{-2pt}
\end{figure}

\paragraph{User study} \Cref{fig:userstudy} shows the human preference of our method versus SDXL. We used standard prompts in PartiPrompt2~\cite{yu2022scaling} to generate 100 images using both methods and ask people for their preference blindly after shuffling. We received a total of three thousand responses on the comparisons of visual quality and text faithfulness, with our method emerging as the clear winner by a clear margin.

\subsection{\ours{}-T2V Results}\label{sec:exp-t2v}

In \Cref{fig:vis_t2v}, We showcase several samples generated by \ours{}-T2V, which are not only visually striking but also temporally coherent. This highlights \ours{}-T2V's effectiveness in creating videos that are both aesthetically pleasing and consistent over time.

\begin{figure}[t]
    \centering
    \includegraphics[width=0.98\linewidth]{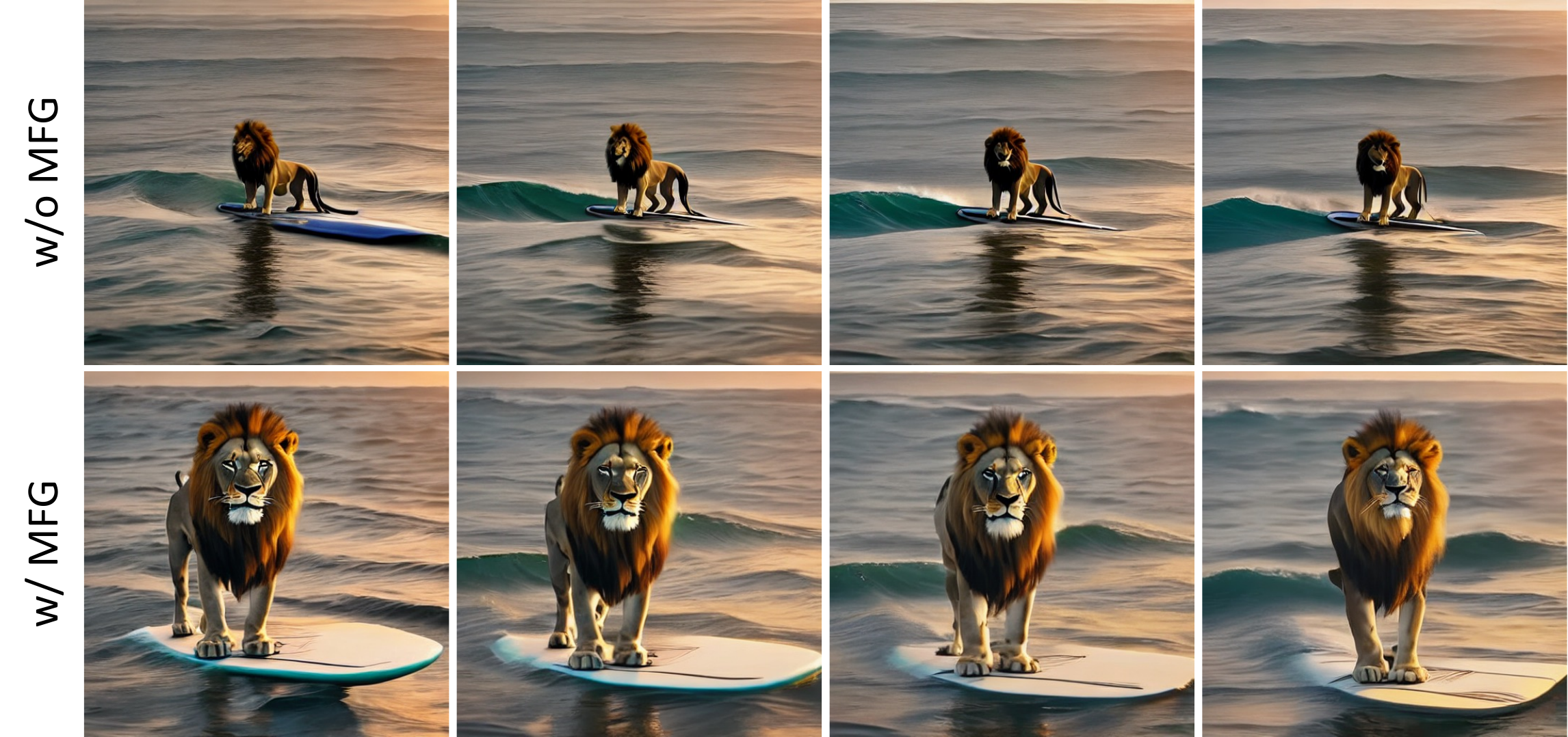}
    \vspace{-8pt}
    \caption{\textbf{Effect of motion-free guidance.} \ours{}-T2V with motion-free guidance has a clear visual appearance improvement. The prompt is {\footnotesize``A lion standing on a surfboard in the ocean in sunset''}.}
    \label{fig:ablation-mfg}
    \vspace{-6pt}
\end{figure}

\paragraph{Effect of motion-free guidance.}
In \Cref{fig:ablation-mfg}, we present a comparison between our \ours{} variants, with motion-free guidance (MFG) and without. For this comparison, critical factors such as the pre-trained T2I model, training data, and the number of training iterations were kept constant to ensure a fair evaluation. The results clearly indicate that \ours{}-T2V, when integrated with MFG, shows a marked tendency to focus on the central object mentioned in the prompt, often rendering it in greater detail. Specifically, the object typically occupies a more prominent, central position in the generated video, thereby dominating the visual focus across video frames.

\section{Conclusion}

In this work, we provide a thorough exploration of transformer-based diffusion models for text-conditioned image and video generation. Our findings shed light on the properties of various conditioning approaches and offer compelling evidence of quality improvement when scaling up the model.
A notable contribution of our work is the development of GenTron for video generation, where we introduce motion-free guidance. This innovative approach has demonstrably enhanced the visual quality of generated videos. 
We hope that our research will contribute to bridging the existing gap in applying transformers to diffusion models and their broader use in other domains.

\vspace{10pt}
\noindent\textbf{Acknowledgement.} This paper is partially supported by the National Key R\&D Program of China No.2022ZD0161000 and the General Research Fund of Hong Kong No.17200622. 

%%%%%%%%% REFERENCES
{\small
\bibliographystyle{ieeenat_fullname}
\bibliography{egbib}
}
\clearpage
\appendix

\section{Summary of Conditioning Mechanism}

\begin{table}[t]
    \small
    \centering
    \setlength{\tabcolsep}{2pt}
    \renewcommand{\arraystretch}{1.1}
    \begin{tabular}{l|c|c}
    Method & Text Encoder & Integration  \\
    \toprule
    \multicolumn{3}{c}{single text encoder} \\
    \hline
    GLIDE~\cite{nichol2021glide} & CLIP-B~\cite{radford2021learning} &  cross-attention \\
    SDv1.4 / SDv1.5 & CLIP-L~\cite{radford2021learning} & cross-attention \\
    SDv2.0 / SDv2.1 & OpenCLIP-H~\cite{openclip} & cross-attention  \\
    DALL-E 2~\cite{ramesh2022hierarchical} & CLIP~\cite{radford2021learning} & cross-attention \\
    DALL-E 3~\cite{dalle3} & T5-XXL~\cite{chung2022scaling} & cross-attention \\
    Imagen~\cite{saharia2022photorealistic} & T5-XXL~\cite{chung2022scaling} &  cross-attention  \\
    DeepFloyd IF~\cite{shonenkov2023deepfloydif} & T5-XXL & cross-attention \\
    DiT~\cite{peebles2022scalable} & N/A & adaLN \\
    PixArt-$\alpha$~\cite{chen2023pixart} & T5-XXL~\cite{chung2022scaling} & cross-attention \\
    \toprule
    \multicolumn{3}{c}{multiole text encoders} \\
    \hline
   eDiff-I~\cite{balaji2022ediffi} & CLIP-L \& T5-XXL & cross-attention\\
   SDXL~\cite{podell2023sdxl} & CLIP-L \& OpenCLIP-bigG & cross-attention \\
   Emu~\cite{dai2023emu} & CLIP-L \& T5-XXL & cross-attention \\
    \end{tabular}
    \caption{Summary of conditioning strategies used by existing image diffusion models.}
    \label{tab:condition_survey}
\end{table}

In \Cref{tab:condition_survey}, we present a summary of the conditioning approaches utilized in existing text-to-image diffusion models. Pioneering studies, such as~\cite{nichol22aglide, ramesh2022hierarchical},  have leveraged CLIP's language model to guide text-based image generation. Furthermore, Saharia~\etal~\cite{saharia2022photorealistic} found that large, generic language models, pretrained solely on text, are adept at encoding text for image generation purposes. Additionally, more recently, there has been an emerging trend towards combining different language models to achieve more comprehensive guidance~\cite{balaji2022ediffi, podell2023sdxl, dai2023emu}. In this work, we use the interleaved cross-attention method for scenarios involving multiple text encoders, while reserving plain cross-attention for cases with a single text encoder. The interleaved cross-attention technique is a specialized adaptation of standard cross-attention, specifically engineered to facilitate the integration of two distinct types of textual embeddings. This method upholds the fundamental structure of traditional cross-attention, yet distinctively alternates between different text embeddings in a sequential order. For example, in one transformer block, our approach might employ CLIP embeddings, and then in the subsequent block, it would switch to using Flan-T5 embeddings.

\begin{figure}[t]
    \centering
    \includegraphics[width=0.96\linewidth]{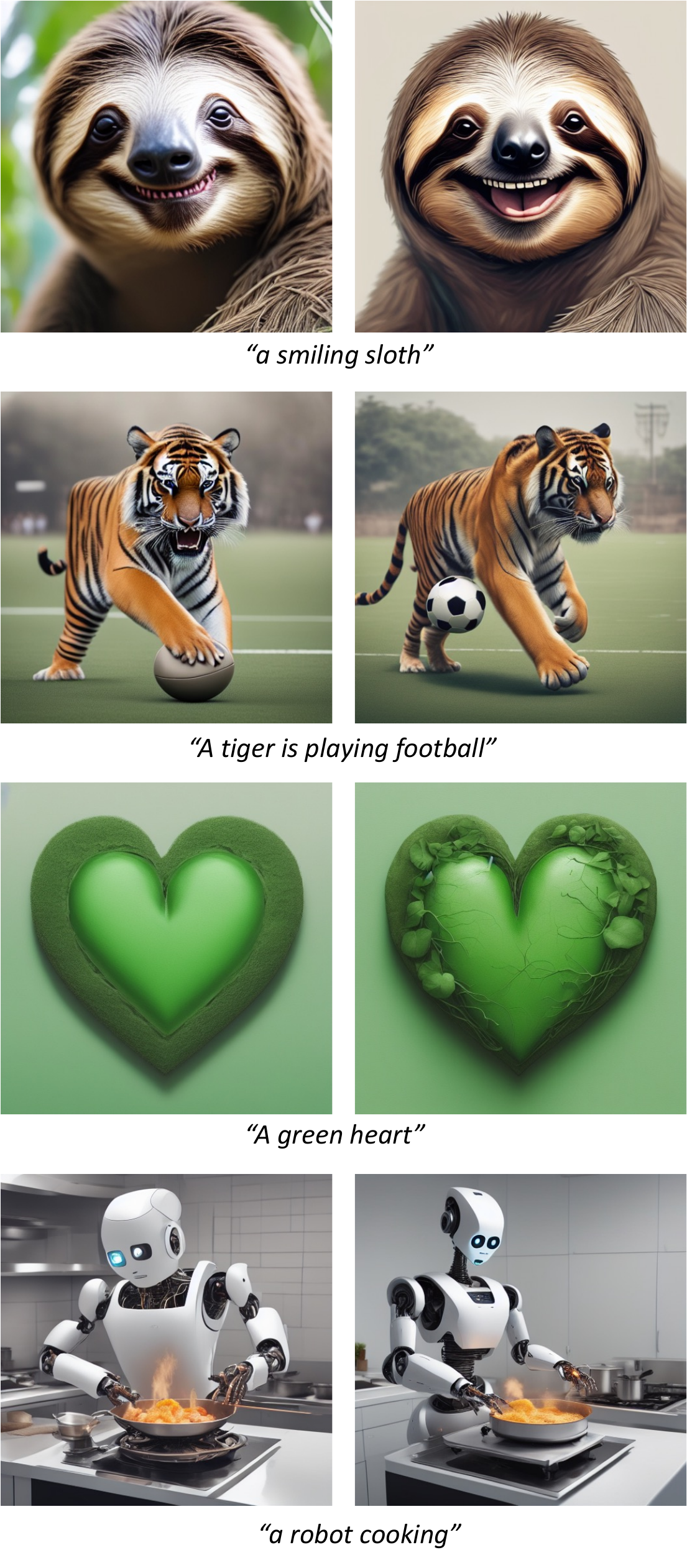}
    \begin{minipage}{0.48\linewidth}
        \centering
        \vspace{-0.6cm}
        \par \ours{}-XL/2
    \end{minipage}
    \begin{minipage}{0.48\linewidth}
        \centering
        \vspace{-0.6cm}
        \par \ours{}-G/2
    \end{minipage}
    \caption{\textbf{More examples of model scaling-up effects.} Both models use the CLIP-T5XXL conditioning strategy. Captions are from PartiPrompt~\cite{yu2022scaling}.}
    \label{fig:scaleup_appendix}
\end{figure}

\section{More Results}

\subsection{Additional Model Scaling-up Examples}
We present additional qualitative results of model scaling up in \Cref{fig:scaleup_appendix}. All prompts are from the PartPrompt~\cite{yu2022scaling}.

\section{Additional \ours{}-T2I Examples}

\begin{figure*}
    \includegraphics[width=0.96\textwidth]{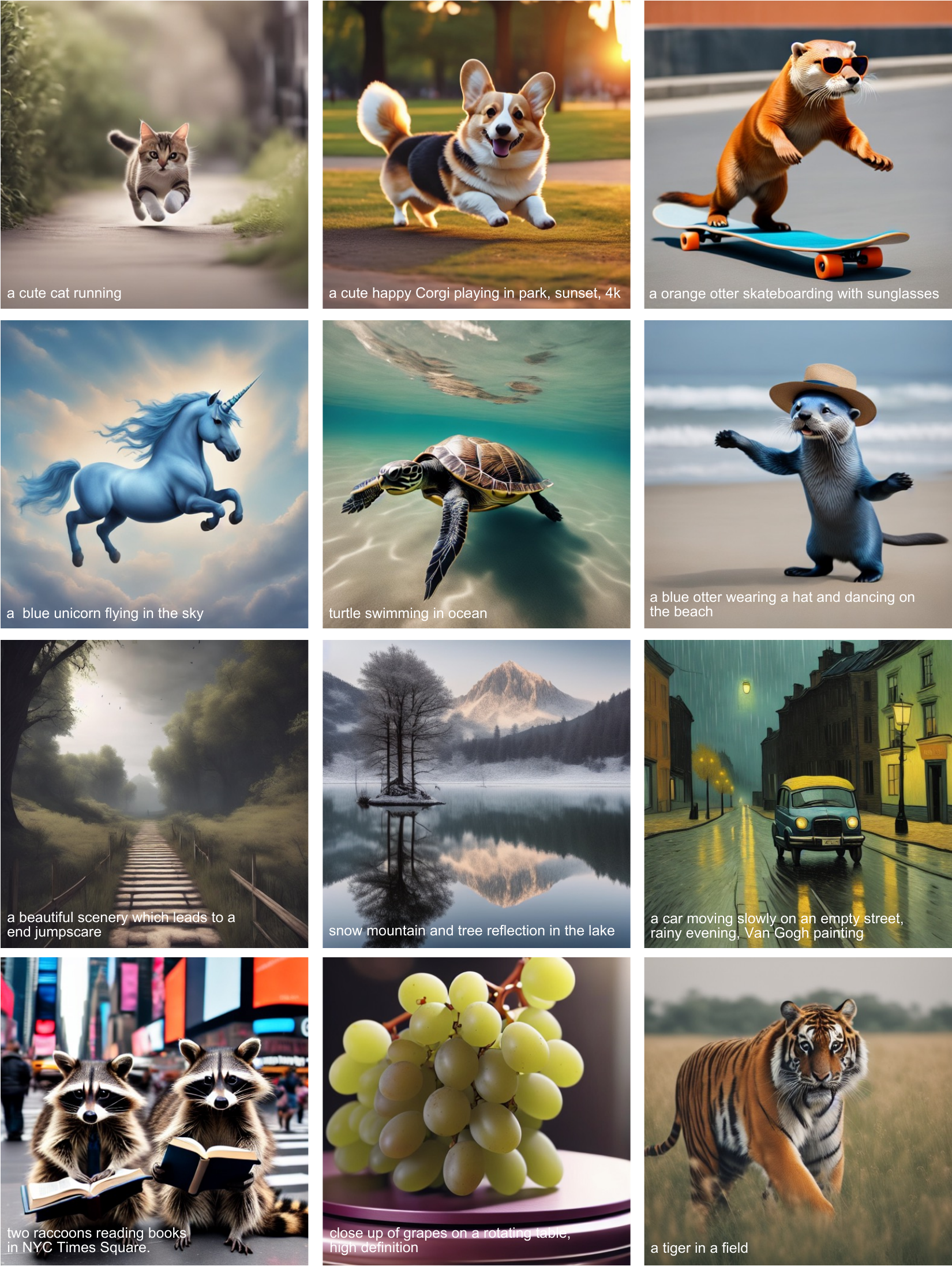}
    \caption{\textbf{\ours{}-T2I examples.} }\label{fig:t2i_demo_appendix}
\end{figure*}

We present more \ours{}-T2I example in \Cref{fig:t2i_demo_appendix}. 

\section{Additional \ours{}-T2V Examples}
Additional GenTron-T2V results are available on our website\textsuperscript{1}.

\end{document}